\documentclass[11pt]{article}

\usepackage{EMNLP2022}

\usepackage{times}
\usepackage{latexsym}

\usepackage{multirow}

\usepackage[T1]{fontenc}

\usepackage[utf8]{inputenc}

\usepackage{microtype}

\usepackage{inconsolata}

\usepackage{float}
\usepackage{graphicx}
\usepackage{subcaption}
\usepackage{amsfonts}
\usepackage{amssymb}
\usepackage{amsmath}

\title{Not Just Plain Text! Fuel Document-Level Relation Extraction with Explicit Syntax Refinement and Subsentence Modeling}

\author{Zhichao Duan\textsuperscript{\rm 1}, 
Xiuxing Li\textsuperscript{\rm 2,3}\Thanks{\hspace{.3em}Corresponding authors.}, 
Zhenyu Li\textsuperscript{\rm 1}, 
Zhuo Wang\textsuperscript{\rm 1}, 
Jianyong Wang\textsuperscript{\rm 1}\footnotemark[1]\\
\textsuperscript{\rm 1}Department of Computer Science and Technology, Tsinghua University\\
\textsuperscript{\rm 2}Key Laboratory of Intelligent Information Processing\\
Institute of Computing Technology, Chinese Academy of Sciences (ICT/CAS)\\
\textsuperscript{\rm 3}University of Chinese Academy of Sciences\\
\texttt{\{dzc20,zy-li21,wang-z18\}@mails.tsinghua.edu.cn}\\
\texttt{lixiuxing@ict.ac.cn}; \texttt{jianyong@tsinghua.edu.cn}
}

\begin{document}
\maketitle
\begin{abstract}
Document-level relation extraction (DocRE) aims to identify semantic labels among entities within a single document. 
One major challenge of DocRE is to dig decisive details regarding a specific entity pair from long text.
However, in many cases, only a fraction of text carries required information, even in the manually labeled supporting evidence.
To better capture and exploit instructive information, we propose a novel exp\textbf{L}icit synt\textbf{A}x \textbf{R}efinement and \textbf{S}ubsentence m\textbf{O}deli\textbf{N}g based framework (LARSON).
By introducing extra syntactic information, LARSON can model subsentences of arbitrary granularity and efficiently screen instructive ones. Moreover, we incorporate refined syntax into text representations which further improves the performance of LARSON.
Experimental results on three benchmark datasets (DocRED, CDR, and GDA) demonstrate that LARSON significantly outperforms existing methods.
\end{abstract}

\section{Introduction}

Relation extraction (RE) is an essential task in information extraction. It aims to model relational patterns between entities in unstructured text. 
One strikingly significant variant of RE, document-level relation extraction (DocRE) is designed to identify relations among entity pairs distributed throughout the document. 
Compared to traditional sentence-level RE (\citealp{dixit-al-onaizan-2019-span,lyu-chen-2021-relation,zhou2021improved}), where entities are located in the same sentence, DocRE further fits the need in real scenes and has received increasing attention lately (\citealp{christopoulou-etal-2019-connecting, wang-etal-2020-global, xu2021document, xie-etal-2022-eider, zhao2022document}).

In an ordinary document, interactions between entities are complex. Since pre-trained language models (PLMs) have shown their great potential in many downstream tasks (\citealp{duan2021bridging,li2022effective}), some works implicitly capture such interactions through pre-trained language models (PLMs) (\citealp{wang2019fine, yuan2021document, zhou2021document, xie-etal-2022-eider}). Some other works model this information explicitly.
They first construct document graphs that consist of different nodes (e.g., mentions, entities, sentences, or the document) (\citealp{christopoulou-etal-2019-connecting, zeng-etal-2020-double, wang-etal-2020-global, xu-etal-2021-discriminative}). Then, graph convolutional networks (GCNs) (\citealp{kipf2016semi, brody2021attentive}) are adopted to encode precise node representations and infer final relations. According to their experiments, GCNs can better capture complex interconnections between nodes which is the foundation of our method. 

\begin{figure}
\centering{\includegraphics[height=0.73\linewidth]{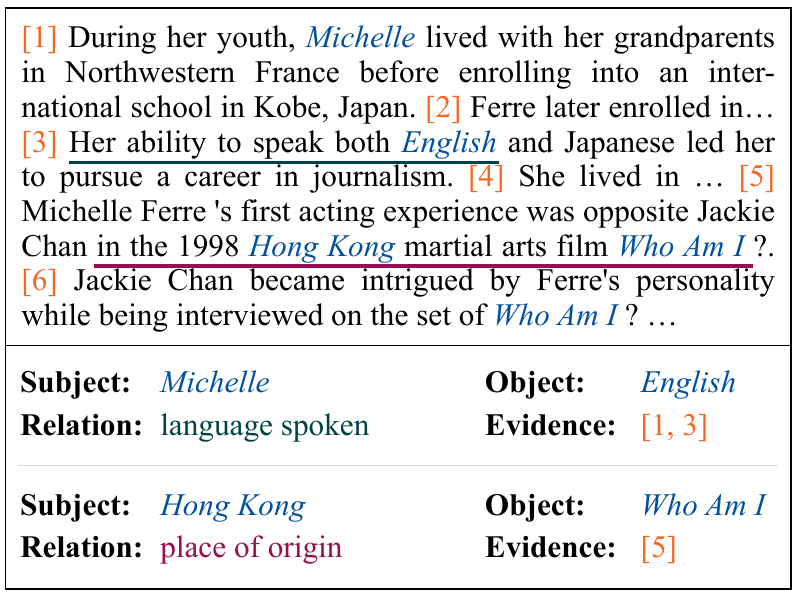}}
\caption{We take an inter-sentence and an intra-sentence relation instance from DocRED.
Decisive subsentences are underlined with different colors according to relation \textcolor[RGB]{0,67,70}{\emph{language spoken}} and \textcolor[RGB]{151,14,83}{\emph{place of origin}}. To identify the relation among (\textcolor[RGB]{0,80,157}{\emph{Michelle}}, \textcolor[RGB]{0,80,157}{\emph{English}}), extra coreference resolution between referring expression \emph{Her} and referent \emph{Michelle} is needed.
}
\label{fig_exam}
\end{figure}

One major challenge around DocRE concerns inferring relations of multiple entity pairs across long text, which may contain irrelevant or even noisy information. 
Take Figure \ref{fig_exam} as an example. It includes an inter-sentence and an intra-sentence relation instance.
To infer the inter-sentence relation between \emph{Michelle} and \emph{English}, human can quickly locate the subsentence "\emph{Her ability ...}" and deduce that \emph{Her} is referring to \emph{Michelle} here.
Furthermore, given (\emph{Hong Kong}, \emph{Who Am I}), human can identify their intra-sentence relation just by a prepositional phrase "\emph{in the 1998 Hong Kong martial arts film Who Am I?} ". 
Unfortunately, such fine-grained annotations are not available in current DocRE datasets.
With massive irrelevant information, DocRE models sometimes struggle with complicated relation instances, indicating that implicitly learning instructive context is not sufficient.
Another factor that hinders the further development of DocRE algorithms is syntax information. As pointed out in \citealp{sundararaman2019syntax,bai-etal-2021-syntax}, though PLMs are trained with massive real-world text data, there is still a big gap between implicitly learned syntax and the golden syntax. Moreover, syntax information is widely incorporated in many sentence-level RE models (\citealp{xu-etal-2016-improved,zhang-etal-2018-graph,qin-etal-2021-relation}) and is not yet developed sufficiently under DocRE scenario (\citealp{gupta2019neural}).

\begin{figure}[!t]
\begin{subfigure}{1.0\columnwidth}
\includegraphics[width=\textwidth]{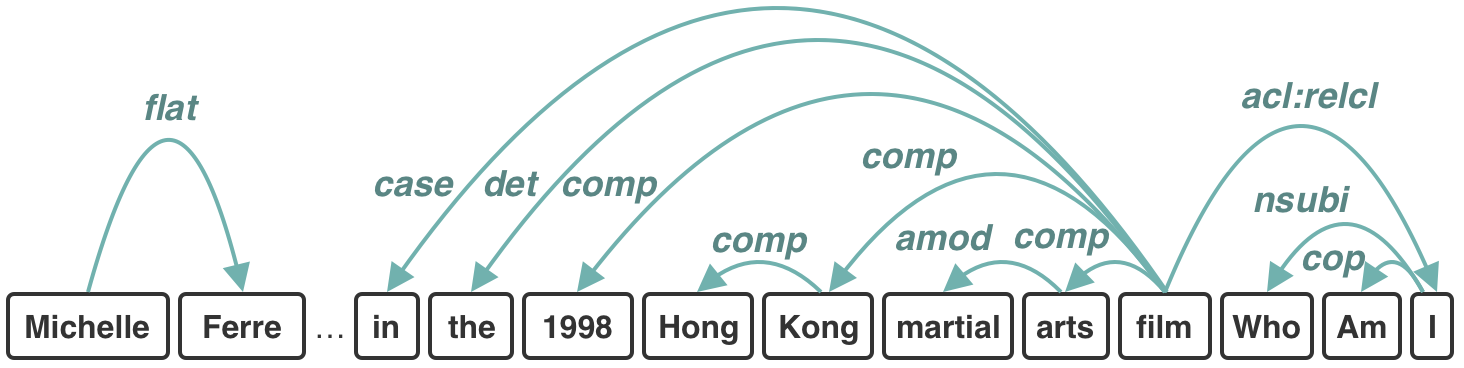}
\caption{Dependency tree describes dependencies between words within a single sentence.
Exploiting such syntax information can significantly complement the original plain text and facilitate coupling neighbor information.
} 
\label{fig_syn_a}
\vspace{0.5cm}
\end{subfigure}
\begin{subfigure}{1.0\columnwidth} 
\includegraphics[width=\textwidth]{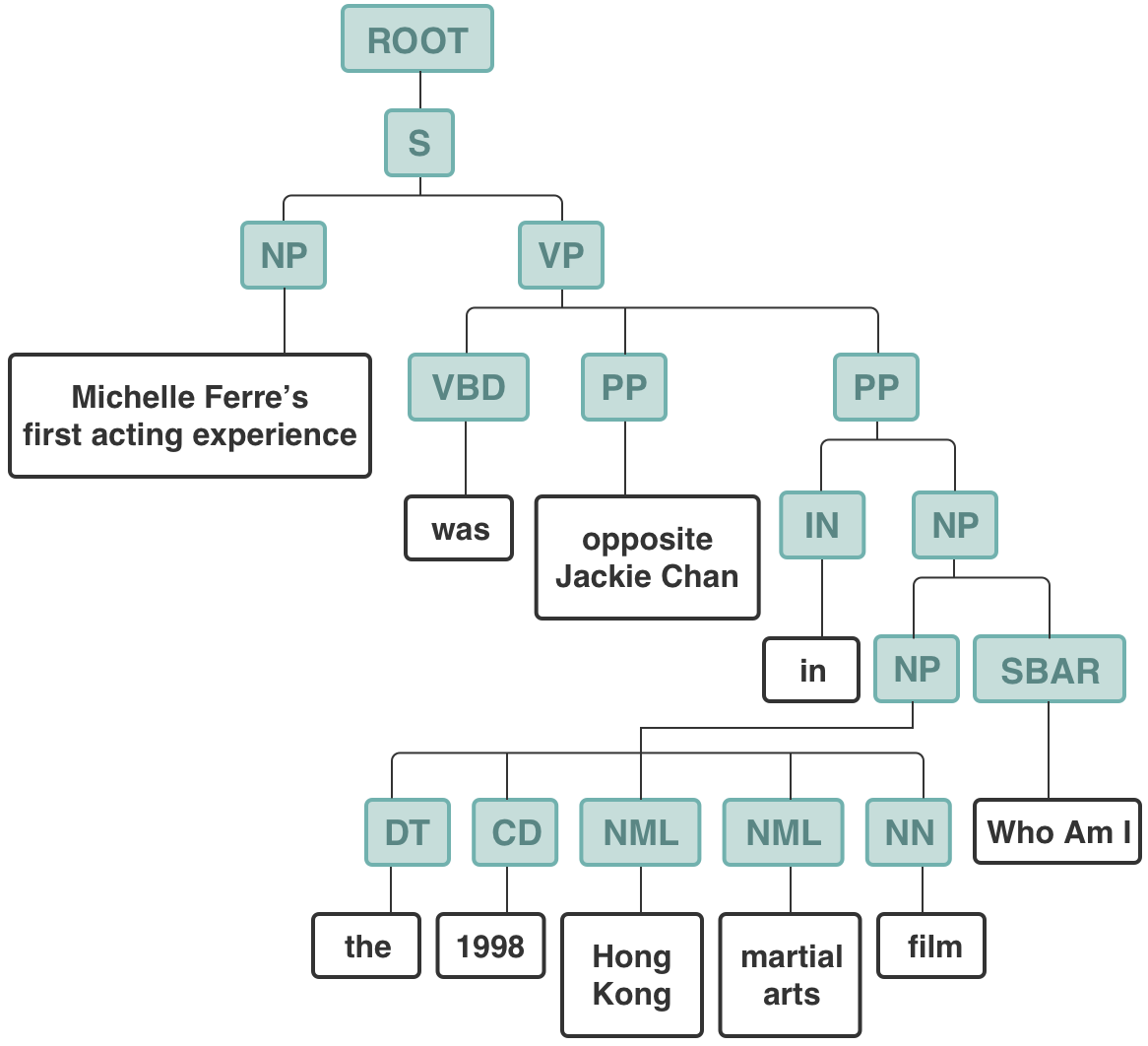} 
\caption{Constituency tree organizes the sentence in a tree structure which not only induces extra hierarchical syntax information but also enables exploring subsentences of arbitrary granularity.}
\label{fig_syn_b}
\end{subfigure}
\caption{Syntactic parsing results of evidence sentence "\emph{Michelle Ferre ...}" mentioned in the previous intra-sentence instance. (a) and (b) represent the corresponding dependency and constituency tree, respectively. Irrelevant nodes (words) are either collapsed or neglected.}
\end{figure}

\begin{figure*}
\centering{\includegraphics[height=0.56\linewidth]{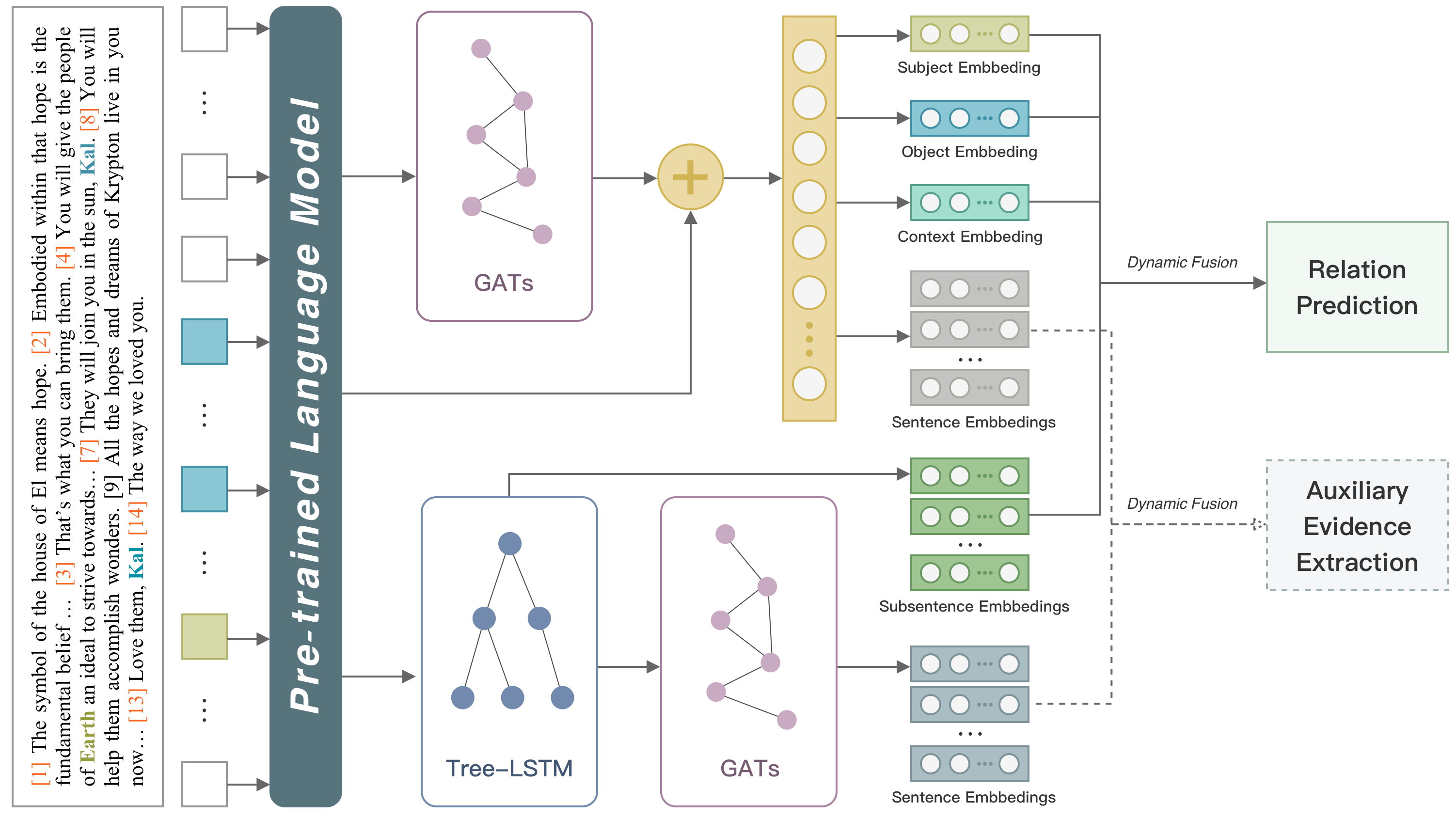}}
\caption{The overall architecture of LARSON. Note that the two GATs share the same architecture but have different parameters.}
\label{overall}
\end{figure*}

In order to better solve DocRE, we propose a novel exp\textbf{L}icit synt\textbf{A}x \textbf{R}efinement and \textbf{S}ubsentence m\textbf{O}deli\textbf{N}g based framework (LARSON). 

We mainly exploit dependency and constituency trees to incorporate extra syntax information and model subsentences.
The two trees depict complete yet different aspects of syntax information.
For example, the evidence sentence as in the previous intra-sentence relation instance can be parsed into corresponding dependency tree and consituency tree\footnote{Dependency and constituency trees are obtained using the Stanza library.\url{https://stanfordnlp.github.io/stanza/}}.
We can see from Figure \ref{fig_syn_a} that a dependency tree describes the dependencies between words within a single sentence that strongly complement the original plain text. Moreover, it facilities coupling syntactically associated words.
The constituency tree shown in Figure \ref{fig_syn_b} can organize different words of a single sentence hierarchically and reasonably, eliminating the process of enumerating different word combinations while maintaining hierarchical syntax information. 
With the aid of dedicated attention module (\citealp{bahdanau2014neural}), we can even gather all subsentences with proper weights to resolve inter-sentence dependencies and produce final relations.

To the best of our knowledge, LARSON is the first DocRE model that considers subsentence modeling. Through extensive experiments on three public DocRE benchmarks, DocRED (\citealp{yao-etal-2019-docred}), CDR (\citealp{li2016biocreative}), and GDA (\citealp{wu2019renet}), we demonstrate that our model LARSON outperforms existing methods by a large margin. 

Our key contributions of this work can be summarized as follows:
\begin{enumerate}
    \item We propose enhancing text representations through dependency trees to complement the original plain text and aggregate information of syntactically associated words.
    \item We propose encoding subsentences through constituency trees to help LARSON focus more on valuable pieces while maintaining hierarchical syntax information.
    \item Experimental results demonstrate that LARSON achieves leading performance on all three DocRE benchmarks, especially on GDA, where our model outperforms state-of-the-art method by 1.48\% \textit{F}$_1$.
\end{enumerate}

\section{Problem Formulation}
Given a document $D$ made up of sentences $\left \{sen_{\,i}\right \}_{i=1}^{I}$ and appearing entities $\left \{e_i\right \}_{i=1}^{N}$, the goal of DocRE is to correctly infer all the existing relations between each entity pair $(e_s,e_o)_{s,o=1...N;s{\neq}o}$ where $e_s$ is the subject entity and $e_o$ is the object entity. 
Among which, entity $e_i$ usually consists of multiple mentions and the $j^{th}$ mention of $e_i$ is denoted as $m_{(i,j)}$. 
Predicted relations are either a subset of predefined relations $\mathcal{R}$ or $\left \{\mathrm{NA}\right \}$ (no relation).

\section{Methodology}

The overall architecture of LARSON is illustrated in Figure \ref{overall}. 
We first extract two kinds of trees and encode plain text using PLMs. With dependency trees and graph attention networks (GATs) (\citealp{brody2021attentive}), syntax complemented text representations are derived.
On this basis, we extract embeddings of subject entity, object entity, and entity pair-related context (Section \ref{text_encoding}). Then, on the strength of Tree-LSTM (\citealp{tai-etal-2015-improved}), we leverage constituency trees to model subsentences of arbitrary granularity (Section \ref{submodel}).
Final relations are inferred based on the dynamic fusion of embeddings obtained in section \ref{text_encoding} and subsentence embeddings obtained in section \ref{submodel}. During the process,
extra sentence embeddings acquired from both trees are utilized for auxiliary evidence extraction to bring into full play of the syntax information (Section \ref{fclass}).

\subsection{Text Encoding}\label{text_encoding}
Given a document $D$, we first insert the marker "*" before and after each mention (\citealp{tacred}). Then, the dependency tree $\mathcal{G}_{dep}$ and constituency tree $\mathcal{G}_{con}$ of each sentence are extracted\footnote{Syntax is parsed in token-level.} using Stanza (\citealp{qi-etal-2020-stanza}). 
We feed tokenized form $\left \{x_i\right \}_{i=1}^{T}$ of $D$ into PLM to obtain contextualized representation $H$ $\in$ $\mathbb{R}^{T{\times}d}$ and token-level attention matrix\footnote{Scores of different heads are averaged.} $A$ $\in$ $\mathbb{R}^{T{\times}T}$ where $T$ is the number of tokens and $d$ is the dimension of token embedding:
\begin{equation}
    H,A = \mathrm{PLM}([x_1,x_2,...,x_T]).
\end{equation}

As pointed out in several studies (\citealp{sundararaman2019syntax,bai-etal-2021-syntax}), there is still a big gap between implicitly learned syntax in PLMs and the golden syntax. In LARSON, we alleviate this issue by first incorporating dependency syntax of each sentence which depicts dependencies between words. 
The contextualized representation $H$ is used as input to the graph attention networks (GATs)  (\citealp{brody2021attentive}) to encode dependency syntax\footnote{Dependency trees are merged by collecting a batch of sentence dependency trees (graphs) and combining them into one graph for efficient graph computation. As a result, each input graph is converted into a disjoint component of the merged graph. During the process, the nodes and edges are relabeled as disjoint segments.} and aggregate information of neighbor nodes:

\begin{equation}
    \mathbf{r}_{j} = {\mathbf{t}}^\intercal \cdot \mathrm{LeakyReLU}\left(
            W^{(a_1)} H_{i}+W^{(a_2)} H_{j}\right)\label{gat1}, \nonumber
\end{equation}
\begin{equation}
    \boldsymbol{\alpha} = \mathrm{softmax}([\mathbf{r}_{1}, \mathbf{r}_{2}, ..., \mathbf{r}_{|\mathcal{N}(i)|}])\label{gat2},
\end{equation}
\begin{equation}
    H_i^{(dep)} = \sum_{j\,\in\,\mathcal{N}(i)} \boldsymbol{\alpha}_{j} W^{(a_2)} H_j\label{gat3}, \nonumber
\end{equation}
where $\boldsymbol{t}$ $\in$ $\mathbb{R}^{d_1}$, $W^{(a_1)}$,$W^{(a_2)}$ $\in$ $\mathbb{R}^{d_1 \times d}$ are trainable parameters. 
Vectors are highlighted in bold. 
$\mathcal{N}(i)$ contains all neighbor nodes that point to node $i$ according to $\mathcal{G}_{dep}$. 
After collecting dependency syntax-aware representation $H^{(dep)}$, we can complement the original text representation $H$ by:
\begin{equation}
    H^{(c)} = H + H^{(dep)}W^{(z)}\label{repadd},
\end{equation}
where $W^{(z)} \in \mathbb{R}^{d_1 \times d}$ is a trainable mapping matrix. 

Then, representation $\mathbf{e}^{(i)}$ of entity $e_i$ is abstracted by merging the embeddings of markers before each associated mention based on logsumexp (\citealp{jia-etal-2019-document}):
\begin{equation}
    {\mathbf{e}}^{(i)} = \mathrm{log} \sum_j \mathrm{exp}({\mathbf{m}}^{(i,\,j)})\label{logsumexp},
\end{equation}
where ${\mathbf{m}}^{(i,j)}$ is the embedding of $m_{(i,j)}$.

In addition to entity-specific embeddings, 
we also extract the localized context embedding (\citealp{zhou2021document}) $\mathbf{c}^{(s,o)}$ to represent the entity pair $(e_s,e_o)$:
\begin{equation}
    {\mathbf{c}}^{(s,o)} = \left( {H^{(c)}} \right)^\intercal \frac{A_s \odot A_o}{A_s^TA_o}\label{attn_context},
\end{equation}
where $A_s, A_o \in \mathbb{R}^T$ represent subject and object entity's attention score to each token in $D$, respectively. ${\odot}$ denotes Hadamard product. Through the above steps, dependency syntax complemented entity and context representations are acquired.
Furthermore, we extract dependency syntax-aware sentence representations by pooling embeddings of all tokens in each sentence using logsumexp (similar to Equation \ref{logsumexp}) based on $H^{(c)}$. 

\subsection{Subsentence Modeling}\label{submodel}
LARSON exploits constituency trees to model subsentences and hierarchical syntax information.
Each constituency tree describes a logical way to restore the entire sentence piece by piece. Utilizing constituency trees, we can not only incorporate extra hierarchical syntax information but also encode subsentences of arbitrary granularity. In practice,
we first set arbitrary node $j$ in a constituency tree with hidden state $\mathbf{n}^{(j)}$ and memory cell state $\mathbf{d}^{(j)}$. Both $\mathbf{n}^{(j)}$ and $\mathbf{d}^{(j)}$ are initialized with zeros.
We set input vectors of all leaf nodes (tokens) with their corresponding representations in the parameter of embedding layer (i.e., embedding matrix $E$) inside PLM. Input vectors for non-leaf nodes are set to zeros.
We then broadcast features of leaf nodes all the way up to the root node by Tree-LSTM (\citealp{miwa-bansal-2016-end}). Input gate $\mathbf{i}^{(j)}$ and output gate $\mathbf{o}^{(j)}$ of arbitrary node $j$ in $\mathcal{G}_{con}$ are calculated as:
\begin{equation}
    \mathbf{i}^{(j)} = \sigma\left(W^{(u)}\mathbf{x}^{(j)}+\sum_{l\in \mathcal{N}(j)}U^{(u)}_l \mathbf{n}^{(j,\,l)} + \mathbf{b}^{(u)}\right), \nonumber
\end{equation}
\begin{equation}
    \mathbf{o}^{(j)} = \sigma\left(W^{(v)}\mathbf{x}^{(j)}+\sum_{l\in \mathcal{N}(j)}U^{(v)}_l \mathbf{n}^{(j,\,l)} + \mathbf{b}^{(v)}\right)\label{lstm1},
\end{equation}
where $\sigma$ is $\mathrm{sigmoid}$ function. $W$,$U$,$\mathbf{b}$ are trainable parameters. $\mathbf{x}^{(j)}$ denotes the input vector of node $j$.
$\mathbf{n}^{(j,\,l)}$ denotes the hidden state of child $l$ of node $j$. 
Integrating input vectors and hidden states is similar to Equation \ref{lstm1}, only trainable parameters are newly initialized and $\mathrm{sigmoid}$ function is replaced with $\mathrm{tanh}$.
Integrated result is marked as $\mathbf{u}^{(j)}$. 

Afterwards, we obtain forget gate of child $k$ of node $j$, i.e., $\mathbf{f}^{(j,\,k)}$ as:
\begin{align}
        \mathbf{f}^{(j,\,k)} = \sigma\left(W^{(f)}\mathbf{x}^{(j)} + \sum_{l\in \mathcal{N}(j)}U_{kl}^{(f)} \mathbf{n}^{(j,\,l)} + \nonumber \right.\\
        \left. \mathbf{b}^{(f)} \right)\label{lstm2}.
\end{align}

At last, we update cell state and hidden state:
\begin{equation}
    \mathbf{d}^{(j)} = \mathbf{i}^{(j)} \odot \mathbf{u}^{(j)} + \sum_{l\in \mathcal{N}(j)} \mathbf{f}^{(j,\,l)} \odot \mathbf{d}^{(j,\,l)}\label{lstm5}, \nonumber
\end{equation}
\begin{equation}
    \mathbf{n}^{(j)} = \mathbf{o}^{(j)} \odot \textrm{tanh}(\mathbf{d}^{(j)})\label{lstm6},
\end{equation}
where $\mathbf{d}^{(j\,,l)}$ denotes the cell state of child $l$ of node $j$. Through Equation \ref{lstm1}-\ref{lstm6}, we can extract hidden states of arbitrary nodes in $\mathcal{G}_{con}$ to represent subsentences. To reduce computation complexity,
only root nodes of subtrees that have at least two leaf nodes are kept to represent subsentences. Since broadcasting process is tightly related to the structure of tree, each node contains hierarchical syntax information. 
In order to further complement sentence representations acquired in section \ref{text_encoding}, we adopt GATs of the same architecture as mentioned before (Equation \ref{gat2}) to merge adjacent node features in constituency trees. Then, hierarchical syntax-aware sentence embeddings can be calculated by averaging hidden states of all nodes inside each tree. Later on, we dynamically fuse acquired subsentence and sentence embeddings with various outputs from section \ref{text_encoding} which will be explained in the next section.

\subsection{Dynamic Fusion and Classification}\label{fclass}
After getting the above representations, we first fuse entity embeddings and context embeddings with subsentence features accordingly using dedicated attention module. 
Given arbitrary embedding $\mathbf{v}\in\mathbb{R}^{d}$ and $B$ subsentence embeddings $[\mathbf{n}^{(1)}, \mathbf{n}^{(2)}, ..., \mathbf{n}^{(B)}]$ where $\mathbf{n}^{(i)} \in \mathbb{R}^{d_1}$, the attention score
$\boldsymbol{\beta} \in \mathbb{R}^B$ is calculated as (\citealp{bahdanau2014neural}):
\begin{equation}
    \mathbf{q}_i = \mathbf{w}^\intercal \cdot \mathrm{tanh}(W^{(b_1)}\mathbf{v}+W^{(b_2)}\mathbf{n}^{(i)})\label{attn1}, \nonumber
\end{equation}
\begin{equation}
    \boldsymbol{\beta} = \mathrm{softmax}([\mathbf{q}_1, \mathbf{q}_2, ..., \mathbf{q}_B])\label{attn2},
\end{equation}
where $\mathbf{w} \in \mathbb{R}^{d_2}$, $W^{(b_1)} \in \mathbb{R}^{{d_2} \times d}$, $W^{(b_2)} \in \mathbb{R}^{{d_2} \times {d_1}}$ are trainable parameters. 
$\boldsymbol{\beta}$ describes the importance of different subsentences with regarding to different components.
We need to emphasize that subsentences of different sentences are all taken into consideration in this step to resolve inter-sentence dependencies.
With the guidance of attention score and a trainable mapping matrix $W^{{(m)}} \in \mathbb{R}^{d \times d_1}$, we can dynamically fuse any embedding $\mathbf{v}$ with subsentence representations using weighted sum:
\begin{equation}
    \mathbf{\hat{v}} = \mathbf{v} + W^{{(m)}} \sum_{i=1}^{B}\boldsymbol{\beta}_i \, \mathbf{n}^{(i)}\label{apply}.
\end{equation}


Through Equation 9-10, we can combine the subject entity embedding, object entity embedding, and context embedding with subsentence representations to form the enhanced representations $\hat{\mathbf{e}}^{(s)}$, $\hat{\mathbf{e}}^{(o)}$, and $\hat{\mathbf{c}}^{(s,o)}$ by replacing $\mathbf{v}$ with $\mathbf{e}^{(s)}$, $\mathbf{e}^{(o)}$, and $\mathbf{c}^{(s,o)}$ respectively. Then, we can calculate the score of relation $r$ (\citealp{zhou2021document}):

\begin{equation}
    \mathbf{z}_s = \mathrm{tanh}(W^{(t_1)}\mathbf{\hat{e}}^{(s)} + W^{(t_2)}\mathbf{\hat{c}}^{(s,\,o)}), \nonumber
\end{equation}
\begin{equation}
    \mathbf{z}_o = \mathrm{tanh}(W^{(q_1)}\mathbf{\hat{e}}^{(o)} + W^{(q_2)}\mathbf{\hat{c}}^{(s,\,o)}),
\end{equation}
\begin{equation}
    l_{\left(r|e_s, e_o\right)} = \mathbf{z}_s^{\,\intercal}W^{(r)}\mathbf{z}_o + b^{(r)}, \nonumber
\end{equation}
where $W$ and $b$ are trainable parameters. 
In order to reach the full potential of refined syntax, we combine constituency syntax-aware sentence embeddings obtained in section \ref{submodel} with dependency syntax-aware sentence embeddings obtained in section \ref{text_encoding} using dedicated attention module (Equation \ref{attn1}-\ref{apply}). With the combined sentence embedding $\mathbf{s}^{(i)}$, 
\begin{table}
\centering
\resizebox{\linewidth}{!}{
\begin{tabular}{lccc} 
\hline
\textbf{Statistics}                & \textbf{DocRED} & \textbf{CDR} & \textbf{GDA}    \\ 
\hline
\# Train                  & 3053   & 500 & 23353  \\
\# Dev                    & 1000   & 500 & 5839   \\
\# Test                   & 1000   & 500 & 1000   \\
\# Relations              & 97     & 2   & 2      \\
Avg.\# sentences per Doc. & 8.0    & 9.7 & 10.2   \\
\hline
\end{tabular}
}
\caption{Statistics of three benchmarks used in our experiments.}
\label{statistics}
\end{table}

\begin{table}
\centering
\resizebox{\linewidth}{!}{
\begin{tabular}{lccc} 
\hline
\textbf{Hyperparam}      & \textbf{DocRED}          & \textbf{CDR}           & \textbf{GDA}            \\
                & BERT & SciBERT & SciBERT  \\ 
\hline
Training epoch  & 30              & 30            & 10             \\
lr for encoder  & 3e-5            & 3e-5          & 2e-5           \\
lr for the rest & 2e-4            & 7e-5          & 5e-5           \\
\hline
\end{tabular}
}
\caption{Hyper-parameters used in three benchmarks.}
\label{hyper}
\end{table}
we can calculate the probability of $sen_{\,i}$ to be an evidence (\citealp{xie-etal-2022-eider}) as:
\begin{equation}
    p_{\left(r_{sen_{\,i}}|e_s, e_o\right)} = \sigma \left( \mathbf{s}^{(i)\,\intercal}W^{(g)}\mathbf{\hat{c}}^{(s,\,o)} + b^{(g)} \right),
\end{equation}
where $W^{(g)}$ and $b^{(g)}$ are trainable parameters.

For relation prediction and evidence extraction of entity pair $(e_s,e_o)$, we adopt adaptive thresholding loss (\citealp{zhou2021document}) (Equation \ref{reloss}) and binary cross entropy loss (Equation \ref{bceloss}) respectively:
\begin{align}
    \mathcal{L}_{RE} & = -\mathrm{log} \left( \frac{\mathrm{exp}(l_{(\mathrm{TH}|e_s,e_o)})}{\sum_{r^\prime \in {\mathcal{N}_T \cup \left\{ \mathrm{TH} \right\}}}  \mathrm{exp}(l_{(r^\prime|e_s,e_o)})} \right) \nonumber \\
    -\sum_{r \in \mathcal{P}_T} & \mathrm{log} \left( \frac{\mathrm{exp}(l_{(r|e_s,e_o)})}{\sum_{r^\prime \in {\mathcal{P}_T \cup \left\{ \mathrm{TH} \right\}}}  \mathrm{exp}(l_{(r^\prime|e_s,e_o)})} \right)\label{reloss},
\end{align}

\begin{align}
    & \mathcal{L}_{Evi} = -\sum_{{sen_{\,i}} \in D}[y_i \cdot \mathrm{log}p_{\left(r_{sen_{\,i}}|e_s, e_o\right)} + \nonumber \\
    & \qquad (1-y_i) \cdot \mathrm{log}(1-p_{\left(r_{sen_{\,i}}|e_s, e_o\right)})]\label{bceloss},
\end{align}
where $\mathcal{P}_T$ and $\mathcal{N}_T$ denote expressed relations and non-expressed relations respectively. $\mathrm{TH}$ is a dummy class introduced to separate positive labels from negative labels. $y_i$ indicates whether sentence $sen_{\,i}$ is an evidence.

The overall loss is defined as a combination of $\mathcal{L}_{RE}$ and $\mathcal{L}_{Evi}$ with constant value $\eta$:
\begin{equation}
    \mathcal{L} = \mathcal{L}_{RE} + \eta \cdot \mathcal{L}_{Evi}\label{allloss}.
\end{equation}

During inference, we follow \citealp{xie-etal-2022-eider} and exploit heuristic rules to construct evidence sentences for relation extraction. For more detailed description, we refer interested readers to it.

\section{Experiments}
\subsection{Datasets} In order to fully evaluate our model, we conduct comprehensive experiments on three widely used public DocRE datasets. Statistics of these datasets are listed in Table \ref{statistics}.
\begin{itemize}
    \item \textbf{DocRED} (\citealp{yao-etal-2019-docred}) is a large-scale crowd-sourced DocRE dataset constructed from Wikipedia articles. It provides 3,053 documents for training covering various domains and requires DocRE models to possess numerous reasoning abilities (e.g., coreference reasoning, or commonsense reasoning).
    \item \textbf{CDR} (\citealp{li2016biocreative}) is a biomedical DocRE dataset built from 1,500 PubMed abstracts which is randomized into three equal parts for training, validating, and testing. It is manually labeled with binary relations between Chemical and Disease concepts.
    \item \textbf{GDA} (\citealp{wu2019renet}) is also a biomedical DocRE dataset contains 30,192 MEDLINE abstracts. The dataset is annotated with binary relations between Gene and Disease concepts using distant supervision. We split GDA according to \citealp{christopoulou-etal-2019-connecting}.
\end{itemize}
\subsection{Implementation Details}
LARSON is implemented based on Pytorch (\citealp{paszke2019pytorch}) and Huggingface's Transformers (\citealp{wolf-etal-2020-transformers}). 
For all experiments, the number of layers in GATs is set to 3 with only 1 attention head. Output dimensions of GATs are 256. Hidden state and cell state of each node in a constituency tree also share dimension 256. We dropout (\citealp{srivastava2014dropout}) attention score inside dedicated attention module with a probability of 50\%. Batch size is set to 4.

Linear learning rate warmup (\citealp{goyal2017accurate}) with ratio 0.06 is deployed followed by a linear decay to 0. $\eta$ in Equation \ref{allloss} is set to 0.1. Entire model is optimized by AdamW optimizer (\citealp{loshchilov2017decoupled}) and tuned on dev set. Mean score of 5 repeated experiments with different random seeds is reported. The rest dataset-specific settings are listed in Table \ref{hyper}.

\begin{table*}
\centering
\resizebox{\linewidth}{!}{
\begin{tabular}{lcccccc} 
\hline
\multirow{2}{*}{\textbf{Model}}      & \multicolumn{4}{c}{\textbf{Dev}}                                               & \multicolumn{2}{c}{\textbf{Test}}       \\ 
\cline{2-7}
                            & \textbf{Ign \textit{F}$_1$} & \textbf{\textit{F}$_1$} & \textbf{Intra \textit{F}$_1$} & \textbf{Inter \textit{F}$_1$} & \textbf{Ign \textit{F}$_1$} & \textbf{\textit{F}$_1$}  \\ 
\hline
\textit{Graph-based Methods}
\\
LSR-BERT$_\mathrm{base}$ (\citealp{nan-etal-2020-reasoning})            & 52.43           & 59.00                & 65.26             & 52.05             & 56.97           & 59.05                 \\
GLRE-BERT$_\mathrm{base}$ (\citealp{wang-etal-2020-global})            & -               & -                    & -                 & -                 & 55.40           & 57.40                 \\
HeterGSAN-BERT$_\mathrm{base}$ (\citealp{xu2021document})       & 58.13           & 60.18                & -                 & -                 & 57.12           & 59.45                 \\
GAIN-BERT$_\mathrm{base}$ (\citealp{zeng-etal-2020-double})            & 59.14           & 61.22                & 67.10             & 53.90             & 59.00           & 61.24                 \\ 
\hline
\textit{PLMs-based Methods}
\\
BERT$_\mathrm{base}$ (\citealp{wang2019fine})                 & -               & 54.16                & 61.61             & 47.15             & -               & 53.20                 \\
BERT-TS$_\mathrm{base}$ (\citealp{wang2019fine})              & -               & 54.42                & 61.80             & 47.28             & -               & 53.92                 \\
HIN-BERT$_\mathrm{base}$ (\citealp{tang2020hin})            & 54.29           & 56.31                & -                 & -                 & 53.70           & 55.60                 \\
CorefBERT$_\mathrm{base}$ (\citealp{ye2020coreferential})            & 55.32           & 57.51                & -                 & -                 & 54.54           & 56.96                 \\
ATLOP-BERT$_\mathrm{base}$ (\citealp{zhou2021document})           & 59.22           & 61.09               & -             & -             & 59.31           & 61.30                 \\
DocuNet-BERT$_\mathrm{base}$ (\citealp{zhang2021document})         & 59.86           & 61.83                & -                 & -                 & 59.93           & 61.86                 \\
EIDER-BERT$_\mathrm{base}$ (\citealp{xie-etal-2022-eider})             & 60.51           & 62.48                & 68.47             & 55.21             & 60.42           & 62.47                 \\ 
\hline
\textbf{LARSON-BERT$_\mathrm{base}$}   & \textbf{61.05}   & \textbf{63.01}      &    \textbf{68.63}               &        \textbf{55.75}            & \textbf{60.71}                &   \textbf{62.83}                    \\ 

\hline
\end{tabular}
}
\caption{Results ($\%$) of relation extraction on the dev and test set of DocRED. 
The best result on test set is reported according to submissions on CodaLab.
Results of other methods are directly taken from original papers.
}
\label{docredres}
\end{table*}

\subsection{Results on DocRED}
We compare LARSON with graph-based methods (\citealp{zeng-etal-2020-double,nan-etal-2020-reasoning,wang-etal-2020-global,xu2021document}) and PLMs-based methods (\citealp{wang2019fine,tang2020hin,ye2020coreferential,zhou2021document,zhang2021document,xie-etal-2022-eider}) on DocRED. We report not only \textit{F}$_1$ and Ign \textit{F}$_1$ (\textit{F}$_1$ score excluding the relational facts shared by the training and dev/test set) as the prior studies (\citealp{yao-etal-2019-docred}), but also Intra \textit{F}$_1$ (\textit{F}$_1$ that only considers intra-sentence relational facts) and Inter \textit{F}$_1$ (\textit{F}$_1$ that only considers inter-sentence relational facts). Experimental results listed in Table \ref{docredres} show that LARSON can achieve leading performance on a general domain DocRE dataset. Specifically, LARSON can improve \textit{F}$_1$ score on dev/test set by 0.53\%/0.36\% over previous state-of-the-art method EIDER (\citealp{xie-etal-2022-eider}). 
The advance confirms that syntax information and subsentence modeling are two crucial factors in DocRE.
More specifically, we can observe performance boosts in both Inter \textit{F}$_1$ (+0.54\%) and Intra \textit{F}$_1$ (+0.16\%), indicating the extensiveness of our method and the efficiency in targeting inter-sentence instances.
\begin{table}
\centering
\resizebox{\linewidth}{!}{
\begin{tabular}{lcc} 
\hline
\textbf{Model}          & \textbf{CDR}   & \textbf{GDA}    \\ 
\hline
LSR-BERT (\citealp{nan-etal-2020-reasoning})       & 64.8  & 82.2   \\
SciBERT (\citealp{zhou2021document})       & 65.1  & 82.5   \\
DHG-BERT (\citealp{zhang-etal-2020-document})      & 65.9  & 83.1   \\
GLRE-SciBERT (\citealp{wang-etal-2020-global})   & 68.5  & -      \\
ATLOP-SciBERT (\citealp{zhou2021document})   & 69.4  & 83.9   \\
EIDER-SciBERT (\citealp{xie-etal-2022-eider})  & 70.63 & 84.54  \\ 
\hline
\textbf{LARSON-SciBERT} & \textbf{71.59} & \textbf{86.02}  \\
\hline
\end{tabular}
}
\caption{Results ($\%$) of relation extraction on test set of CDR and GDA. We choose the best checkpoint based on dev set to evaluate the final performance.
Result of SciBERT is based on the re-implemented version (\citealp{zhou2021document}). Other results are directly taken from original papers.
}
\label{biores}
\end{table}

\begin{table}[!t]
\centering
\resizebox{\linewidth}{!}{
\begin{tabular}{lcccc} 
\hline
\textbf{Ablation}              & \textbf{Ign \textit{F}$_1$} & \textbf{\textit{F}$_1$} & \textbf{Intra \textit{F}$_1$} & \textbf{Inter \textit{F}$_1$}  \\ 
\hline
LARSON-BERT            & \textbf{61.05}   & \textbf{63.01}      &    68.63               &        \textbf{55.75}         \\
w/o dependency tree   & 60.77      & 62.73  & 68.62        & 55.22         \\
w/o constituency tree & 60.54      & 62.60  & \textbf{68.69}        & 55.02         \\
w/o dynamic fusion    & 60.24      & 62.46  & 68.36        & 55.20         \\
\hline
\end{tabular}
}
\caption{Ablation study of LARSON on dev set of DocRED.}
\label{ablation}
\end{table}

\subsection{Results on biomedical Datasets}
Besides general domain DocRE dataset DocRED, we also compare LARSON with various advanced methods (\citealp{nan-etal-2020-reasoning,zhang-etal-2020-document,wang-etal-2020-global,zhou2021document,xie-etal-2022-eider}) on two biomedical domain datasets CDR and GDA. Experimental results are listed in Table \ref{biores}. In summary, LARSON achieves significant improvements over two tested datasets (+0.96\% \textit{F}$_1$ on CDR and +1.48\% \textit{F}$_1$ on GDA). 
As we can see in Table \ref{statistics}, CDR and GDA have more sentences in a document on average compared to DocRED. In these complex documents, LARSON manages to skillfully conduct relation extraction, proving the reasonableness and capability of designed architecture. The fact that LARSON can work well in biomedical domain further demonstrates its generality.

\subsection{Ablation Study}
To exhaustively understand how each component contributes to final performance, we conduct three ablation studies and list the results in Table \ref{ablation}. \textit{w/o dependency tree} removes dependency syntax encoding module. All output embeddings in section \ref{text_encoding} are calculated based on plain output of PLM. We can observe that without dependency syntax, decline in {Inter \textit{F}$_1$} (-0.53\%) is much more obvious than it in {Intra \textit{F}$_1$} (-0.01\%). 
Similar trends happen when we remove constituency syntax and subsentence modeling as in the \textit{w/o constituency tree}. Inter \textit{F}$_1$ decreases to 55.02\% which is even worse than \textit{w/o dependency tree}.
Surprisingly consistent results prove that both dependency and constituency syntax are crucial when it comes to inter-sentence instances. 
Noting that while the overall performance drops in \textit{w/o constituency tree}, {Intra \textit{F}$_1$} uncommonly increases. It suggests there is still room for improving how constituency information is integrated, especially in the intra-sentence scenario.
\textit{w/o dynamic fusion} removes dynamic fusing of output embeddings in section \ref{text_encoding} and section \ref{submodel}. Instead, we calculate the average of different subsentence embeddings and directly add them to entity/context embeddings after being mapped to proper shapes. Dependency syntax-aware and constituency syntax-aware sentence embeddings are directly added together in a sentence-wise manner to carry out auxiliary task.
Without dynamic fusion, the performance of LARSON reduces dramatically in all aspects which fully demonstrates the necessity of this module.
Different subsentences are associated with different entities. Brutally combining them together using equal weights neglects the interconnections between the two components. 

\subsection{Case Study}
\begin{figure}
\centering{\includegraphics[width=1\linewidth]{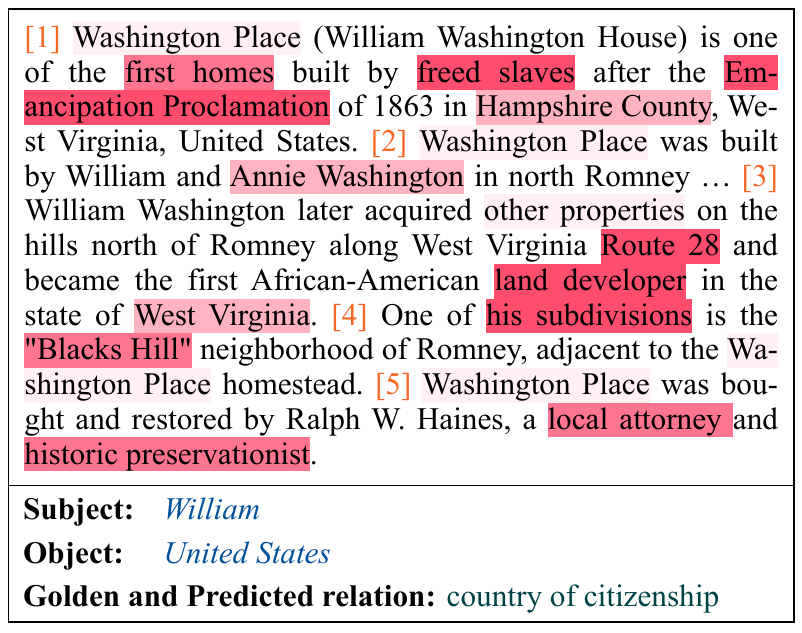}}
\caption{Attention weights of an example in DocRED with regarding to entity pair (\textcolor[RGB]{0,80,157}{\emph{William}}, \textcolor[RGB]{0,80,157}{\emph{United States}}).}
\label{fig_case}
\end{figure}

To intuitively demonstrate that LARSON can link entity pairs with instructive subsentences, we take one example in dev set of DocRED and visualize it in Figure \ref{fig_case}. 
Twenty subsentences with the highest attention scores are colored pink in different depths. Similar ones like \emph{Proclamation} and \emph{Emancipation Proclamation} are merged.
As shown in Figure \ref{fig_case}, LARSON assigns extremely high relevance score to event \textit{Emancipation Proclamation}, role \textit{freed slaves}, place \textit{Route 28}, role \textit{land developer}, and other characters \textit{his subdivisions}.
Among which, human can infer the relation between \emph{William} and \emph{United States} by a key subsentence "\emph{African-American land developer}" in the 3$^{rd}$ sentence. The fact that LARSON also views \emph{land developer} as a highly relevant subsentence indicates our model can efficiently capture decisive details. 
Besides that, all marked subsentences have direct or indirect connections with at least one of subject and object. This example is a strong proof that with subsentence modeling, LARSON can filter out inconsequential pieces and better concentrate on meaningful information.

\section{Related Work}
Relation extraction (RE) is an indispensable part of many real-world applications like question answering (\citealp{hixon-etal-2015-learning}), biomedical text analysis (\citealp{hong2020novel}), etc. Many studies have been done to solve sentence-level RE where entities are within the same sentence, such as sequence-based methods (\citealp{zeng-etal-2014-relation, tacred}), graph-based methods (\citealp{miwa-bansal-2016-end, wu2019simplifying}), and PLMs based methods (\citealp{han2021ptr, zhou2021improved}). \citealp{nguyen-moschitti-2011-joint} even successfully utilized both dependency and constituency trees to better solve sentence-level RE task.
These methods have shown tremendous performance in sentence-level RE task. Nevertheless, according to \citealp{yao-etal-2019-docred}, a considerable proportion of relation instances can only be inferred from multiple sentences, and existing methods are still inadequate for real-world RE problems. DocRE, due to its more realistic setting, has gained more and more attention lately. With the rapid development of PLMs, they have been widely adopted as the first step in DocRE to encode plain text. 
According to the rest steps, existing high performance methods can be divided into two categories. One category inherits implicitly learned attention distribution inside PLMs or designs extra attention modules to capture long range dependencies. \citealp{yuan2021document} uses PLMs to encode semantic features of document and weights different sentences based on different entities via attention mechanism. In order to combine sentence-level and document-level features together, gating mechanism is developed. Instead, \citealp{zhou2021document} inherits implicitly learned attention distribution inside PLMs to determine entity-related context. In addition, they also adopt adaptive threshold to better distinguish positive labels from negative labels. 
The other category exploits GCNs to capture complex interactions between different components (e.g., entities, or sentences) and conduct logical reasoning. \citealp{zeng-etal-2020-double} proposes a heterogeneous mention-level graph to capture complex interactions among different mentions. They further aggregate different mention representations to construct an entity-level graph. Based on the graphs, they develop a novel path reasoning mechanism for final relation extraction.
\citealp{xu2021document} encourages the model to reconstruct reasoning paths while identifying correct relations. 
Unlike previous works, LARSON tries to better solve DocRE task by integrating explicitly refined syntax and subsentence modeling. To the best of our knowledge, LARSON is the first work investigating the effects of subsentences under the DocRE scenario.

\section{Conclusion}

In this work, we propose a novel LARSON model for document-level relation extraction task. LARSON mostly exploits two kinds of extra syntax information, namely dependency syntax and constituency syntax. Graph attention networks and Tree-LSTM are adopted to encode the two kinds of information. Furthermore, through leveraging dedicated attention module, we can dynamically weight different subsentences to assist LARSON in capturing instructive information regardless of the granularity. Experiments on three public DocRE datasets demonstrate that our LARSON model outperforms existing methods by a large margin.

\section*{Limitations}
The challenge to extract instructive information not only exists in DocRE but also in many other document-level tasks (e.g., reading comprehension, document retrieval). For now, our hypothesis is merely tested in DocRE.

\section*{Acknowledgements}
We thank Ziyue Wang (Tsinghua University) for designing all the figures used in this work. We also thank Zhengyan Zhang (Tsinghua University) for 
his valuable and constructive suggestions during the planning and development of this work and his detailed comments on the manuscript.
This work was supported in part by National Key Research and Development Program of China under Grant No. 2020YFA0804503, National Natural Science Foundation of China under Grant No. 62272264, and Beijing Academy of Artificial Intelligence (BAAI).

\bibliography{emnlp2022}
\bibliographystyle{emnlp2022}




\end{document}